\documentclass[unnumsec,webpdf,contemporary,large]{oup-authoring-template} %

\usepackage{graphicx}
\usepackage{booktabs}
\usepackage{bbm}
\usepackage{hyperref}
\usepackage[final]{changes}
\usepackage{setspace}

\graphicspath{{Fig/}}

\theoremstyle{thmstyleone}%
\theoremstyle{thmstyletwo}%
\theoremstyle{thmstylethree}%

\begin{document}

\journaltitle{Journal of the American Medical Informatics Association}
\DOI{DOI HERE}
\copyrightyear{2022}
\pubyear{2019}
\access{Advance Access Publication Date: Day Month Year}
\appnotes{Paper}

\firstpage{1}


\title{Let Large Language Models Judge Each Other: Multi-Agent Peer-Reviewed Reasoning for Medical Question Answering}

\author[1]{Zaifu Zhan, MEng}
\author[2]{Shuang Zhou, PhD}
\author[2,*]{Rui Zhang, PhD}

\authormark{Zaifu Zhan et al.}

\address[1]{\orgdiv{Department of Electrical and Computer Engineering}, 
            \orgname{University of Minnesota}, 
            \orgaddress{\street{200 Union St SE}, \postcode{55455}, \state{Minneapolis, MN}, \country{United States}}}

\address[2]{\orgdiv{Division of Computational Health Sciences, Department of Surgery}, 
            \orgname{University of Minnesota}, 
            \orgaddress{\street{420 Delaware St SE}, \postcode{55455}, \state{Minneapolis, MN}, \country{United States}}}

\corresp[$\ast$]{Corresponding author:
Dr. Rui Zhang, PhD, Division of Computational Health Sciences, Department of Surgery, University of Minnesota, Office: D528 Mayo building, 420 Delaware St SE, Minneapolis, MN 55455,
\href{mailto:zhan1386@umn.edu}{zhan1386@umn.edu}, Office Phone: 612-626-4209\\
\\
Full paper word count: 3306\\
Abstract word count: 193
}

\abstract{
\doublespacing
\textbf{Objective}:
To enhance the accuracy, interpretability, and robustness of large language models (LLMs) in medical question answering (MedQA).\\
\textbf{Method}:
We designed a multi-agent peer-reviewed reasoning method in which multiple LLM agents independently generate chain-of-thought reasoning with candidate answers, then act as peer reviewers to evaluate each other’s reasoning for factual correctness and logical soundness. The highest-rated reasoning chain is selected to produce the final answer. Experiments were conducted with five state-of-the-art LLMs (Llama-3.1-8B, Qwen2.5-7B, Phi-4, DeepSeek-LLM-7B, GPT-oss-20B) on three benchmark datasets: HeadQA, MedQA-USMLE, and PubMedQA. Performance was compared against single-model chain-of-thought reasoning and chain-of-thought-based majority voting.\\
\textbf{Results}:
Peer-reviewed reasoning consistently outperformed both baselines. The best model combination achieved an average accuracy of 0.820 across datasets, exceeding the strongest single model (0.777) and majority voting ensembles (up to 0.789). The method also scaled effectively with more participating models, while peer assessments reliably distinguished high- from low-quality reasoning chains.\\
\textbf{Conclusion}:
The proposed multi-agent peer-reviewed reasoning method enables LLMs to act as both solvers and evaluators, yielding superior performance in MedQA. By emphasizing reasoning quality rather than answer agreement alone, this approach improves accuracy, interpretability, and robustness, offering a promising direction for trustworthy biomedical AI systems.
}



\maketitle
\doublespacing

\section{Introduction}

Large language models (LLMs) have advanced biomedical natural language processing~\cite{chen2025benchmarking,zhou2025large,zhan2025evaluation}. They have shown strong performance in tasks such as named entity recognition~\cite{li2024benchmarking}, relation extraction~\cite{zhan2025ramie}, summarization~\cite{tang2023evaluating}, disease classification~\cite{zhan2025retrieval}, diagnosis~\cite{zhou2025uncertainty}, and question answering~\cite{zhou2025automating}. As model size has increased, LLMs have demonstrated emergent reasoning abilities~\cite{lucas2024reasoning,yang2024llm} that allow them to comprehend complex texts, understand human questions, and even address tasks that were not explicitly learned during training~\cite{zhang2024instruction}. These advances have produced clear improvements in medical question answering (MedQA). The ability to answer questions accurately is particularly important in biomedicine because factual or logical errors may mislead clinical decision-making or research conclusions~\cite{singhal2025toward,jin2022biomedical}. Moreover, most biomedical information needs can be formulated as question–answering problems, which makes improvement in MedQA performance broadly valuable.

\begin{figure*}[htbp]
    \centering
    \includegraphics[width=1\linewidth]{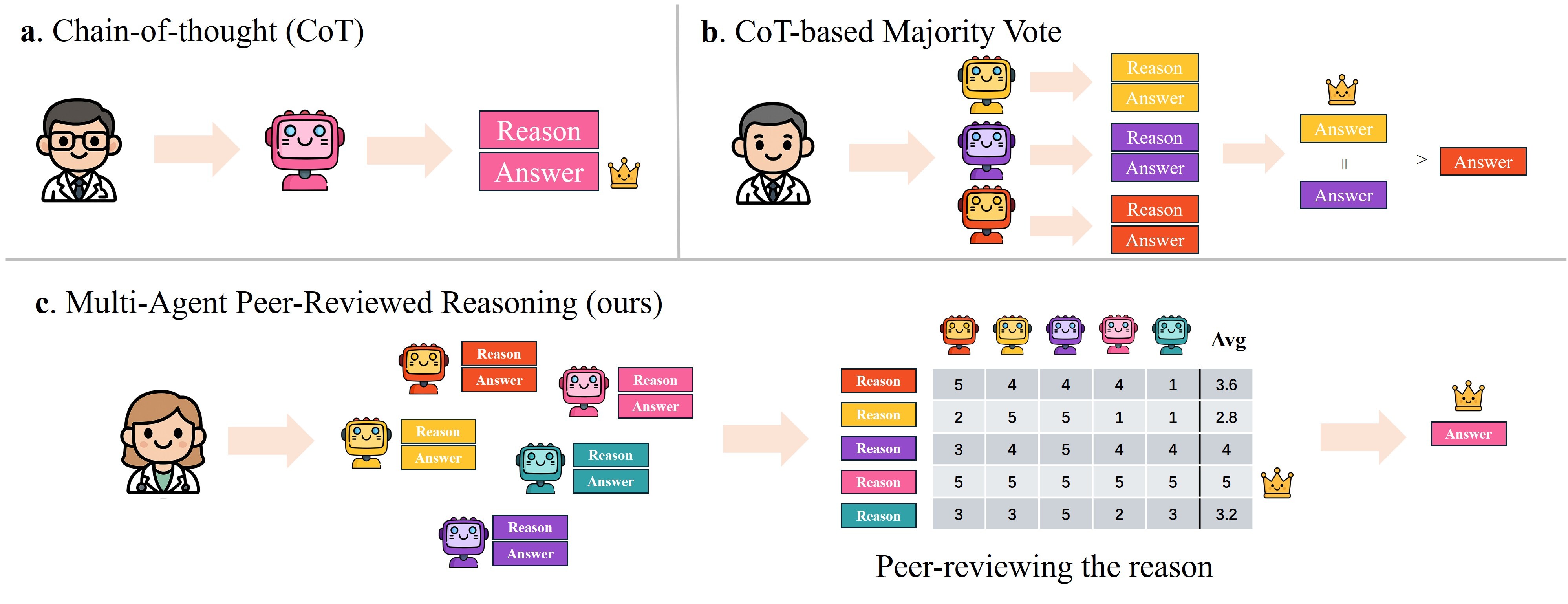}
    \caption{Illustration of three reasoning paradigms. 
    (a) Chain-of-thought (COT): each model independently generates a reasoning process and an answer. 
    (b) COT-based group voting: multiple models produce reasoning–answer pairs, and the final answer is decided by majority voting over the answers. 
    (c) Multi-agent Peer-Reviewed Reasoning (ours): models not only generate reasoning–answer pairs but also evaluate each other’s reasoning; the final answer is selected based on peer review of reasoning quality.}
    \label{fig:method}
\end{figure*}
Recent work has explored various approaches to improving LLMs on MedQA, with growing emphasis on leveraging their reasoning capabilities~\cite{jin2024rjua,zuo2025medxpertqa,lucas2024reasoning}. Researchers have become increasingly interested in whether models can mimic human-like reasoning~\cite{che2025towards}. One landmark development is chain-of-thought (CoT) prompting~\cite{wei2022chain}, which encourages the model to generate answers step by step and often leads to better performance~\cite{che2025towards,liu2024era,jin2024prollm}. Many studies have demonstrated that CoT improves LLM accuracy on QA tasks~\cite{jeon2025comparative,singhal2025toward}. Further progress has been achieved by combining CoT with techniques such as instruction fine-tuning~\cite{le2025instruction}, few-shot prompting~\cite{nachane2024few}, and retrieval-augmented generation~\cite{wang2025medcot}. These findings suggest that reasoning-based paradigms, especially CoT, are becoming a foundation for improving LLM performance on complex QA problems.

For the application of CoT in MedQA, the quality of reasoning is crucial. Correct reasoning usually produces correct answers, while flawed reasoning often leads to wrong ones~\cite{wang2025medcot,lievin2024can,wu2025chain}. Most existing works~\cite{wu2025medcasereasoning,moell2025medical,zhou2025automating} evaluate reasoning after the answer has been produced, either through human inspection or by comparing to human rationales, rather than incorporating reasoning evaluation into the decision process itself.

Two research directions suggest possible solutions to this limitation. The first is LLM-as-a-judge\cite{zheng2023judging}, which has demonstrated that LLMs can evaluate output quality with a high level of agreement with human raters. This approach has been successfully applied to biomedical relation extraction\cite{laskar2025improving}, clinical trial recommendation~\cite{curran2024examining}, and patient record summarization~\cite{croxford2025automating}.
For example, Zhou et al~\cite{zhou2025automating} adopted LLM as a judge and proposed LLM-w-Ref, a novel evaluation framework that leverages fine-grained rationales and LLM-as-a-Judge mechanisms to assess intermediate reasoning with expert-level fidelity while maintaining scalability. 
However, using LLMs as automatic judges has important limitations: their preferences do not always align with human experts,~\cite{wang2023aligning,zheng2023judging} they can overvalue fluent but incorrect answers~\cite{kocmi2023large}, and their scores are sensitive to prompt design and evaluation setup~\cite{arabzadeh2025human}

The second direction is multi-agent collaboration\cite{wang2025survey}, which shows that groups of LLMs can achieve stronger reasoning abilities through interaction. For example, MDAgents\cite{kim2024mdagents} enables multiple LLMs to collaborate, achieving improvements on real-world medical knowledge and clinical diagnosis benchmarks. Similarly, Chen et al.~\cite{chen2025mdteamgpt} designed a multi-agent multi-disciplinary team framework to enhance medical consultation, thereby improving both diagnostic rationality and accuracy.

Inspired by the advantages of LLM-as-a-judge and multi-agent collaboration, we propose a multi-agent peer-review reasoning method to improve MedQA, as illustrated in Figure \ref{fig:method}. In this approach, multiple independent LLM agents generate chain-of-thought solutions to the same medical question. Each agent then reviews the reasoning generated by its peers and rates the logical soundness. The system aggregates these peer reviews to identify the most credible and coherent reasoning chain, which is subsequently used to generate the final answer. Unlike approaches that rely on majority voting over final answers, our method evaluates and selects the reasoning process itself. This enables the production of answers that are more accurate, interpretable, and robust in biomedical contexts.

Our main contributions can be summarized as follows:
\begin{enumerate}
    \item We propose a novel multi-agent peer-review reasoning method that leverages the evaluative and collaborative capacities of LLMs to enhance MedQA.
    \item We conducted extensive experiments, demonstrating that our method is effective across different datasets, models, and agent configurations.
\end{enumerate}

\section{Methods}
\subsection{Overview of methods}
We propose a multi-agent peer-reviewed reasoning method to improve medical question answering. The method consists of two phases: (1) multiple LLM agents independently generate chain-of-thought reasoning with candidate answers, and (2) agents act as peer reviewers, scoring each other’s reasoning for factual correctness and logical soundness. The highest-rated reasoning chain is selected, and its answer is adopted as the final output.

We evaluated this method on five state-of-the-art LLMs (Llama-3.1-8B, Qwen2.5-7B, Phi-
4, DeepSeek-LLM-7B, and GPT-oss-20B) across three benchmark datasets(HeadQA, MedQA-USMLE, and PubMedQA), and compared it against two baselines: single-model chain-of-thought reasoning and CoT-based majority voting. This design allows us to assess both the effectiveness and robustness of our approach.

\subsection{MedQA task and datasets}
To evaluate our proposed multi-agent peer-review reasoning method, we adopt the MedQA task, which serves as a widely recognized benchmark for assessing the reasoning ability of large language models in the biomedical domain. In this task, models are required to answer domain-specific questions, often in multiple-choice format, while demonstrating coherent reasoning chains that reflect both factual knowledge and logical consistency.
We consider three representative datasets: HeadQA, MedQA(USMLE), and PubMedQA. 

HeadQA~\cite{headqa} is constructed from official Spanish healthcare professional exams, covering diverse domains such as medicine, nursing, psychology, and pharmacy. The questions are concise but knowledge-intensive, demanding specialized expertise for accurate answers. 

MedQA(USMLE)~\cite{jin2021disease}, on the other hand, is derived from the United States Medical Licensing Examination and is widely regarded as one of the most challenging benchmarks in biomedical NLP. It consists of high-quality, expert-level multiple-choice questions designed to evaluate not only factual recall but also complex clinical reasoning and diagnostic decision-making. 

Complementary to these exam-style datasets, 
PubMedQA~\cite{jin2019pubmedqa} is built from biomedical research articles in PubMed, where the task is to answer yes/no/maybe research questions based on scientific abstracts. Unlike HeadQA and MedQA, which test professional exam knowledge, PubMedQA emphasizes evidence-based reasoning and the ability to interpret biomedical literature.

Together, these datasets provide a comprehensive and complementary evaluation setting: HeadQA highlights the breadth of domain knowledge, MedQA-USMLE represents the depth of expert-level reasoning, and PubMedQA emphasizes literature-driven inference. This diversity enables us to rigorously assess both the robustness and generalization capacity of our method across different forms of biomedical reasoning.
All the dataset statistics are summarized in Table \ref{tab:medqa_datasets}.
\begin{table}[t]
\centering
\begin{tabular}{lccccc}
\hline
\textbf{Dataset} & \textbf{Testset} & \textbf{Source}\\
\hline
MedQA (USMLE) & 1,273 & Medical exam (USMLE) \\
HeadQA        & 244   & Spanish medical exams \\
PubMedQA      & 500 & Biomedical literature  \\
\hline
\end{tabular}
\caption{Statistics and characteristics of medical QA datasets.}
\label{tab:medqa_datasets}
\end{table}

\subsection{Multi-Agent Peer-Review Reasoning Method}
Inspired by the peer-review process of academic papers—where peer-reviewed work is deemed worthy of publication—we adopt a similar idea in our method. Reasoning and answers that undergo peer review are regarded as high-quality responses.
Thus, we proposed the multi-agent peer-review reasoning method.
By comparison, figure~\ref{fig:method} presents three paradigms: (a) \emph{Chain-of-Thought}, where a single model reasons step-by-step before answering; (b) \emph{Majority Vote}, where multiple models output answers and the most frequent answer is selected; and (c) \emph{Multi-agent peer-review method (this work)}. While majority voting may improve robustness, agreement does not guarantee correctness—models can share the same biases or training artifacts—and it discards valuable reasoning traces. Our method preserves and evaluates reasoning: multiple models first generate CoT traces, then act as judges to score one another’s reasoning. The highest-rated (peer-reviewed) reasoning and its associated answer are selected as the final output.
By combining explicit reasoning generation with cross-model peer review, our method evaluates both performance and the reliability of the underlying reasoning, providing a more informative and trustworthy approach to medical question answering than CoT or majority vote alone.

\subsubsection{Phase 1: Chain-of-Thought Generation}
We elicit explicit, structured reasoning from multiple LLMs before revealing final answers. For multiple-choice medical QA (e.g., MedQA, HeadQA), prompts require option-by-option analysis with stepwise justification~\cite{nachane2024shot,guo2025structured}. For literature-grounded QA (e.g., PubMedQA), prompts instruct models to examine the abstract, extract supporting evidence, and form a logically consistent conclusion~\cite{wang2025medcot}. This phase (i) makes decision paths transparent, (ii) typically improves answer quality via systematic analysis, and (iii) produces rich intermediate traces that are amenable to evaluation.
The CoT prompt could be found in supplementary material 1.

\subsubsection{Phase 2: Peer-Review of Reasoning}
LLMs then evaluate the responses produced by \emph{other} models using a 6-point rubric (0–5) that jointly assesses answer correctness and reasoning quality. Scores of 4–5 denote correct answers with sound medical reasoning; 3 denotes correct answers with minor flaws; 1–2 capture substantial reasoning issues or incorrect answers; 0 reflects fundamental misunderstanding. Importantly, judges do \emph{not} access ground-truth labels; they rely on domain knowledge and internal reasoning, yielding an assessment of \emph{understanding} rather than pattern matching. For each candidate response, the judge receives the original question, the full reasoning trace, and the final answer, and returns a numeric score plus a concise justification. Cross-model evaluations form a score matrix; we aggregate scores (e.g., by mean or median) to select the top peer-reviewed reasoning and adopt its answer as final.
The judgement prompt could be found in supplementary material 1.

\noindent
\textbf{Scoring rubric:} The 0–5 scoring rubric is not intended to replicate formal peer-review scoring standards, but rather serves as an interpretable and lightweight proxy for reasoning quality assessment. Alternative prompting inspirations—such as checklist-based evaluation, pairwise preference judgments, or binary accept/reject schemes—could plausibly influence both model judgments and downstream performance. We leave a systematic comparison of such evaluation strategies to future work.

\noindent
\textbf{Tie handling}: During score aggregation, we observed that ties in the peer-reviewed scores occurred in approximately 15\% of the cases. In such situations, we adopted a simple and deterministic tie-handling strategy: when multiple reasoning chains received identical aggregated scores, the system selected the response that appeared first in the initial generation order. This behavior follows the stability property of Python’s sorting function, which preserves the original relative order of elements with equal keys. We applied the same tie-handling strategy consistently to the CoT-based majority voting baseline, where ties in vote counts were resolved by selecting the first encountered answer.

\subsection{Experiments}
We designed two baselines to evaluate the effectiveness of our method: (1) Chain-of-Thought (CoT) reasoning, where individual models generate reasoning and answers independently, and (2) Majority Voting, where the final answer is determined by aggregating the outputs of multiple models. These baselines provide a comparison for understanding the contributions of reasoning generation and peer review evaluation.

We conducted experiments using five state-of-the-art language models: Llama-3.1-8B~\cite{dubey2024llama}, Qwen2.5-7B~\cite{qwen2,qwen2.5}, Phi-4~\cite{abdin2024phi}, DeepSeek-LLM-7B~\cite{bi2024deepseek}, and GPT-oss-20B~\cite{gptoss}. 
These models were selected to introduce diversity along several concrete axes. 
First, they span a wide range of parameter scales (7B–20B), covering both smaller and larger LLMs in our evaluation pool. 
Second, they come from distinct model families (Llama, Qwen, Phi, DeepSeek, and GPT-oss), which reflect different development lineages and training paradigms. 
Third, this diversity is empirically reflected in heterogeneous baseline performance: the best single model (GPT-oss-20B) achieves an average accuracy of 0.777, whereas DeepSeek-LLM-7B yields 0.437, indicating substantial variation in their reasoning and answering behaviors even before peer review. 
Together, these differences provide a more challenging and informative setting to test whether cross-model peer evaluation can leverage complementary strengths rather than relying on redundant agents.

To evaluate model performance, we used standard accuracy metrics for final answer correctness, following prior works~\cite{headqa,jin2019pubmedqa,jin2021disease}. Additionally, we introduced a peer review scoring system based on a 6-point rubric (0-5) to assess the quality of reasoning. This dual-metric approach captures both the correctness of answers and the logical consistency of reasoning processes.
We report 95\% confidence intervals for accuracy using bootstrap resampling \added{(1000 iterations)} over the test set. For each iteration, accuracy was recomputed on the resampled data, and the 2.5th and 97.5th percentiles were used to define the confidence interval.

All experiments were conducted on NVIDIA A100 GPUs with 40 GB of memory. We used a temperature of 0.1 for less randomness, a maximum generation length of 1,024 tokens, and fixed random seeds for reproducibility. The experiments were implemented using Python, with key libraries including PyTorch~\cite{paszke2019pytorch} and Hugging Face Transformers~\cite{wolf2020transformers}.
In experiments, we decoupled the experiments in three stages: first generating all LLM responses for every question in every dataset, then generating all LLM peer reviews in a separate batch, and finally aggregating the outputs to compute the final performance metrics.
\begin{table*}[htbp]
\caption{Performance comparison of individual LLMs, COT-based group voting, and peer-reviewed reasoning on three medical QA datasets (HeadQA, MedQA, PubMedQA). 95\% confidence intervals and the average accuracy are reported. Throughout the CoT-based majority voting and peer-review reasoning settings, we anonymize model identities by replacing names with numeric indices. The correspondence between indices and models is detailed in the Chain-of-Thought part.
}
\centering
\begin{tabular}{l l c c c c}
\toprule
Method & Model & HeadQA(95\%CI) & MedQA(95\%CI) & PubMedQA(95\%CI) & Avg. \\
\midrule
\multirow{5}{*}{Chain-of-thought} & (1) Deepseek-llm-7b  & 0.4303 (0.3689, 0.4918) & 0.3103 (0.2852, 0.3362) & 0.57 (0.526, 0.614) & 0.437 \\
 & (2) Qwen2.5-7B  & 0.709 (0.6516, 0.7664) & 0.5695 (0.542, 0.597) & 0.67 (0.628, 0.71) & 0.650 \\
 & (3) Llama-3.1-8B  & 0.7213 (0.6639, 0.7746) & 0.6575 (0.6316, 0.6834) & 0.754 (0.716, 0.792) & 0.711 \\
 & (4) Phi-4  & 0.8197 (0.7705, 0.8648) & 0.7313 (0.707, 0.7557) & 0.736 (0.696, 0.774) & 0.762 \\
 & (5) GPT-oss-20B  & 0.8361 (0.7869, 0.8811) & 0.7903 (0.7675, 0.8123) & 0.706 (0.666, 0.746) & 0.777 \\
\midrule
\multirow{15}{*}{COT-based group voting} & (1) + (2) + (3) + (5) & 0.7787 (0.7254, 0.8279) & 0.6897 (0.6638, 0.7148) & 0.698 (0.658, 0.738) & 0.722 \\
 & (1) + (3) + (4) & 0.7787 (0.7254, 0.8279) & 0.6874 (0.6614, 0.7125) & 0.718 (0.678, 0.758) & 0.728 \\
 & (3) + (4) & 0.7951 (0.7418, 0.8443) & 0.6999 (0.6748, 0.7251) & 0.702 (0.662, 0.742) & 0.732 \\
 & (1) + (2) + (4) + (5) & 0.7951 (0.7418, 0.8443) & 0.7266 (0.7023, 0.751) & 0.69 (0.65, 0.73) & 0.737 \\
 & (2) + (3) + (4) & 0.7992 (0.7459, 0.8484) & 0.7015 (0.6764, 0.7266) & 0.728 (0.688, 0.766) & 0.743 \\
 & (1) + (3) + (5) & 0.7869 (0.7336, 0.8361) & 0.7258 (0.7015, 0.7502) & 0.72 (0.68, 0.758) & 0.744 \\
 & (3) + (5) & 0.8033 (0.7541, 0.8525) & 0.7628 (0.7392, 0.7855) & 0.672 (0.63, 0.712) & 0.746 \\
 & (1) + (2) + (3) + (4) + (5) & 0.8115 (0.7623, 0.8607) & 0.7188 (0.6936, 0.7431) & 0.732 (0.692, 0.77) & 0.754 \\
 & (2) + (3) + (5) & 0.8115 (0.7623, 0.8607) & 0.7353 (0.7109, 0.7596) & 0.73 (0.69, 0.768) & 0.759 \\
 & (1) + (3) + (4) + (5) & 0.8115 (0.7623, 0.8607) & 0.7526 (0.729, 0.7761) & 0.72 (0.68, 0.758) & 0.761 \\
 & (1) + (4) + (5) & 0.8156 (0.7664, 0.8648) & 0.7612 (0.7376, 0.7848) & 0.714 (0.674, 0.754) & 0.764 \\
 & (2) + (3) + (4) + (5) & 0.8197 (0.7705, 0.8648) & 0.7502 (0.7258, 0.7738) & 0.726 (0.686, 0.764) & 0.765 \\
 & (4) + (5) & 0.8402 (0.791, 0.8852) & 0.7777 (0.7549, 0.8005) & 0.682 (0.64, 0.722) & 0.767 \\
 & (2) + (4) + (5) & 0.8279 (0.7787, 0.873) & 0.7777 (0.7549, 0.8005) & 0.726 (0.686, 0.764) & 0.777 \\
 & (3) + (4) + (5) & 0.8443 (0.7992, 0.8893) & 0.7832 (0.7604, 0.806) & 0.736 (0.696, 0.774) & 0.788 \\
\midrule
\multirow{15}{*}{Peer-reviewd reasoning} & (2) + (5) & 0.832 (0.7828, 0.877) & 0.7997 (0.7777, 0.8217) & 0.716 (0.676, 0.756) & 0.783 \\
 & (1) + (5) & 0.8361 (0.7869, 0.8811) & 0.8044 (0.7824, 0.8256) & 0.73 (0.69, 0.768) & 0.790 \\
 & (1) + (2) + (5) & 0.8361 (0.7869, 0.8811) & 0.8044 (0.7824, 0.8256) & 0.734 (0.694, 0.772) & 0.791 \\
 & (2) + (3) + (5) & 0.8361 (0.7869, 0.8811) & 0.8052 (0.7832, 0.8264) & 0.744 (0.706, 0.782) & 0.795 \\
 & (1) + (2) + (3) + (5) & 0.8361 (0.7869, 0.8811) & 0.8099 (0.7879, 0.8311) & 0.752 (0.714, 0.79) & 0.799 \\
 & (3) + (5) & 0.8566 (0.8115, 0.8975) & 0.8162 (0.795, 0.8374) & 0.744 (0.706, 0.782) & 0.806 \\
 & (2) + (4) + (5) & 0.8566 (0.8115, 0.8975) & 0.8068 (0.7848, 0.828) & 0.754 (0.716, 0.792) & 0.806 \\
 & (1) + (3) + (5) & 0.8484 (0.8033, 0.8934) & 0.8201 (0.7989, 0.8413) & 0.754 (0.716, 0.792) & 0.807 \\
 & (1) + (2) + (4) + (5) & 0.8566 (0.8115, 0.8975) & 0.8138 (0.7918, 0.835) & 0.756 (0.718, 0.794) & 0.809 \\
 & (4) + (5) & 0.8566 (0.8115, 0.8975) & 0.8225 (0.8013, 0.8429) & 0.752 (0.714, 0.79) & 0.810 \\
 & (2) + (3) + (4) + (5) & 0.8525 (0.8074, 0.8934) & 0.813 (0.791, 0.8342) & 0.772 (0.734, 0.808) & 0.812 \\
 & (1) + (4) + (5) & 0.8607 (0.8156, 0.9016) & 0.8233 (0.802, 0.8437) & 0.754 (0.716, 0.792) & 0.813 \\
 & (1) + (2) + (3) + (4) + (5) & 0.8525 (0.8074, 0.8934) & 0.817 (0.7958, 0.8382) & 0.774 (0.736, 0.81) & 0.814 \\
 & (3) + (4) + (5) & 0.8484 (0.8033, 0.8934) & 0.8264 (0.8052, 0.8468) & 0.774 (0.736, 0.81) & 0.816 \\
 & (1) + (3) + (4) + (5) & 0.8566 (0.8115, 0.8975) & 0.8272 (0.806, 0.8476) & 0.776 (0.738, 0.812) & 0.820 \\
\bottomrule
\end{tabular}
\label{tab:mainres}
\end{table*}
\begin{figure}[t]
    \centering
    \includegraphics[width=1\linewidth]{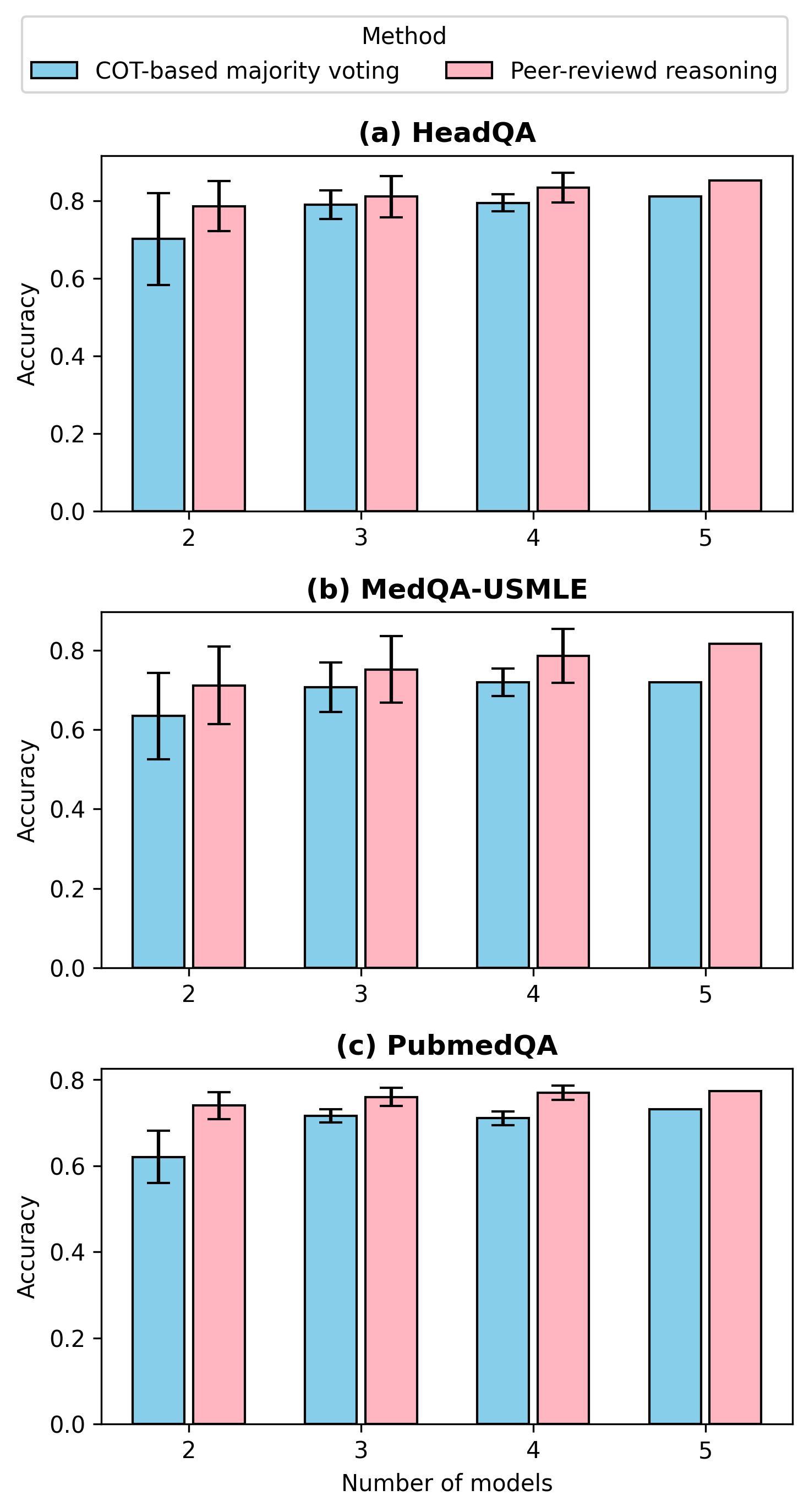}
    \caption{Comparison of COT-based majority voting (blue) and Peer-reviewed reasoning (pink) across different numbers of models (2–5). Panels (a–c) present results on three different 3 datasets. The y-axis shows accuracy with error bars indicating standard deviations. Overall, peer-reviewed reasoning consistently outperforms COT-based majority voting, and performance generally improves as the number of models increases. When the number of models is 5, only one set of experimental data is available, so there is no standard deviation.}
    \label{fig:num_models}
\end{figure}

\section{Results}
As shown in Table \ref{tab:mainres}, we evaluated the performance of the proposed peer-reviewed reasoning method across three representative biomedical QA datasets: HeadQA, MedQA-USMLE, and PubMedQA. Results were compared against two baselines: chain-of-thought reasoning and CoT-based majority voting.
In each method of experiments, we swept through all possible model combinations, ranging from pairwise combinations (10 groups) to triplets (10 groups), quadruplets (5 groups), and quintuplets (1 group). We then selected and sorted the combination with the highest average accuracy for presentation.
\begin{figure*}[htbp]
    \centering
    \includegraphics[width=1\linewidth]{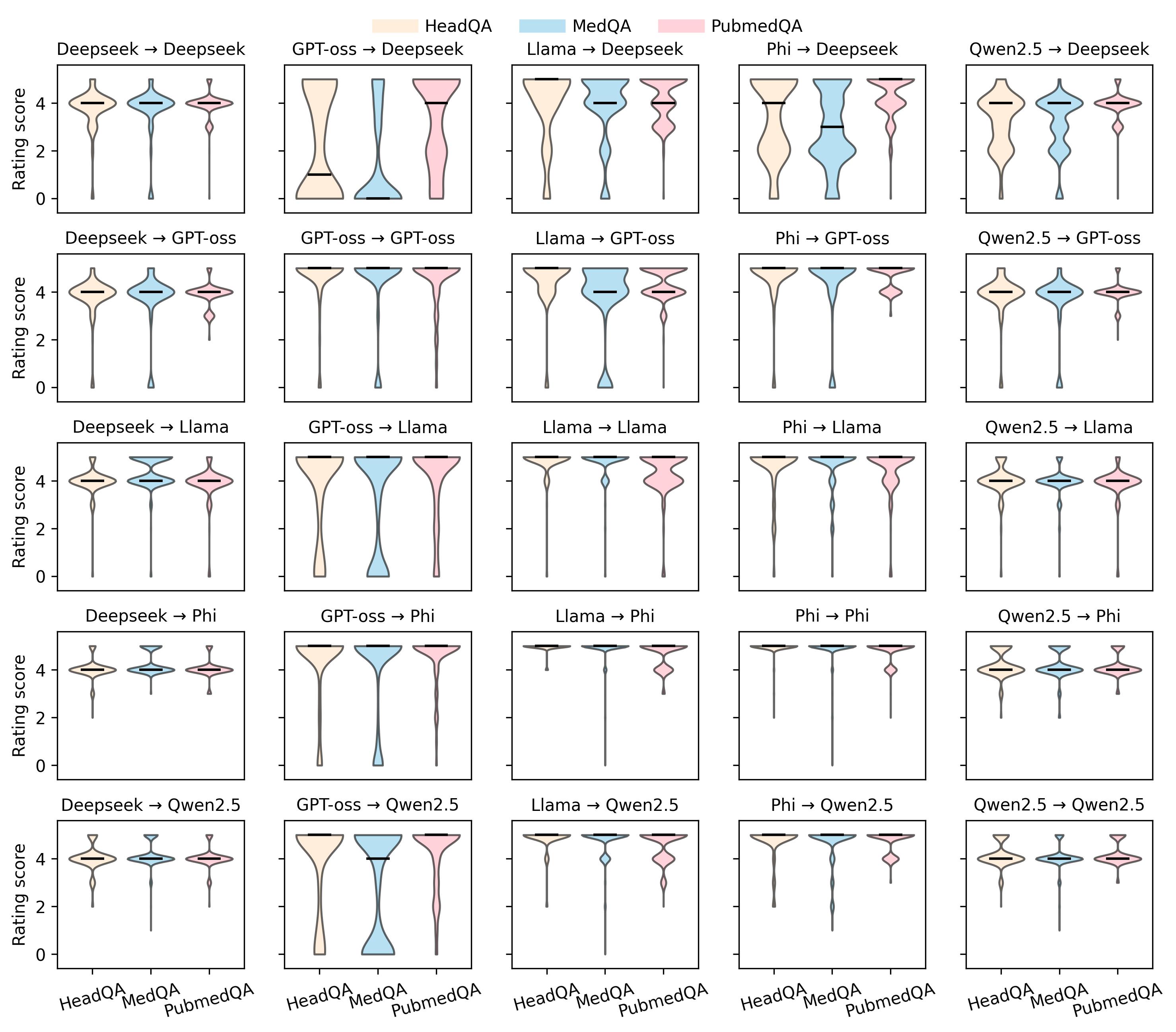}
    \caption{Pairwise rating score distributions across models and datasets. Each subplot corresponds to one judge model (column) evaluating another model (row). Model1 $\rightarrow$ model2 means model1 judges model2's reasoning. Within each subplot, violin plots illustrate the distribution of rating scores on three datasets (HeadQA, MedQA, and PubmedQA), distinguished by different fill colors. The black horizontal line inside each violin indicates the median score.}
    \label{fig:rating_scores}
\end{figure*}

\subsection{Performance of Individual Models.}
Individual LLMs displayed heterogeneous performance across datasets, as shown in Table \ref{tab:mainres}. The best single model, GPT-oss-20B, reached an average accuracy of 0.777, followed by Phi-4 (0.762) and Llama-3.1-8B (0.710). In contrast, DeepSeek-LLM-7B yielded a substantially lower average accuracy of 0.437. These results indicate the variability in baseline performance among different model architectures/sizes and training paradigms.

\subsection{Effect of Majority Voting.}
CoT-based majority voting led to consistent improvements over single-model reasoning. For instance, combining models (3 + 4 + 5) achieved an average accuracy of 0.789, outperforming all individual models. When more models were aggregated, performance generally improved, with accuracies ranging from 0.722 to 0.789 depending on the combination.
In contrast to the performance of an individual model, the aggregated results from multiple models exhibit reduced variance.
These results demonstrate that simple ensemble strategies enhance robustness, though gains remained moderate relative to the best single models.

\subsection{Peer-Reviewed Reasoning results}
Peer-reviewed reasoning consistently outperformed both individual models and majority voting. Across all datasets, accuracy improvements were observed regardless of the specific model combinations. For example, using models (Llama, Phi, GPT-oss) achieved an average accuracy of 0.816, while the combination (Deepseek, Llama, Phi, GPT-oss) yielded the strongest results with an accuracy of 0.820. 
Although DeepSeek exhibits relatively weaker performance as a standalone solver, its inclusion contributes additional diversity in the peer-review stage, which improves the selection of high-quality reasoning chains.
On individual datasets, peer-reviewed reasoning reached up to 0.861 on HeadQA, 0.827 on MedQA-USMLE, and 0.776 on PubMedQA, surpassing the corresponding results from CoT-based majority voting or individual LLM.

\subsection{Scaling with Model Numbers.}
Figure \ref{fig:num_models} illustrates the effect of increasing the number of models on performance. Peer-reviewed reasoning consistently exceeded majority voting across two-, three-, four-, and five-model settings. Accuracy gains were most pronounced in settings with larger model pools, highlighting the scalability of the proposed approach.

\subsection{Reasoning Quality Assessment.}
Beyond accuracy, the peer-review process successfully distinguished reasoning quality across models. As shown in Figure \ref{fig:rating_scores}, pairwise rating distributions reveal that LLM judges provided differentiated assessments of their peers’ reasoning. High-performing models such as GPT-oss-20B (mean:4.1, std:1.2) and Phi-4 (mean:4.4, std:0.9) generally received higher ratings, while weaker models like DeepSeek-LLM-7B (mean:3.2, std:1.6) were consistently rated lower. These patterns confirm that the peer-review mechanism is effective in identifying logically sound reasoning chains, which in turn contribute to improved final answers.
Additionally, to make the relationship between reasoning quality assessment and performance explicit, we computed correlations between the mean peer-review score of each model and its single-model accuracy. We observed a strong positive Pearson correlation (r = 0.91, p = 0.034), indicating that models receiving higher peer-evaluated reasoning scores tend to achieve higher answer accuracy.

Taken together, these findings demonstrate that peer-reviewed reasoning achieves superior performance compared to both chain-of-thought and majority voting baselines. By incorporating reasoning evaluation into the answer selection process, our method yields more accurate, robust, and high-quality outputs across diverse biomedical QA benchmarks.

\subsection{Ablation study: Judge Prompt}


\added{Compared with majority voting, our method introduces an additional mutual judging stage. A key component of this stage is the design of the judge prompt. As shown in Table \ref{tab:prompt_acc}, we compared three different judge prompts on the MedQA-USMLE dataset across 15 model combinations. For each combination, we evaluate three judge prompts (Prompt1–3) and report the mean accuracy as well as the standard deviation across the three prompts.
It is important to note that these results are obtained by evaluating each configuration on the entire test set, rather than through repeated sampling. Therefore, the reported standard deviation reflects the performance variation across different prompt formulations, rather than uncertainty from sampling.}

\added{We observe that the variation across prompts is small, with the largest standard deviation being 0.007. This suggests that the performance of our framework is relatively consistent under different judge prompt designs, indicating a degree of robustness to prompt variations. The detailed designs of the different judge prompts can be found in Supplementary Material 1.}

\begin{table*}[htbp]
\centering
\begin{tabular}{l c c c c c}
\toprule
Model & Prompt1 & Prompt2 & Prompt3 & Avg & Std \\
\midrule
(3)+(4)+(5) & 0.826 & 0.835 & 0.826 & 0.829 & 0.005 \\
(1)+(3)+(4)+(5) & 0.827 & 0.834 & 0.820 & 0.827 & 0.007 \\
(3)+(5) & 0.816 & 0.828 & 0.821 & 0.822 & 0.006 \\
(1)+(4)+(5) & 0.823 & 0.822 & 0.818 & 0.821 & 0.003 \\
(1)+(3)+(5) & 0.820 & 0.828 & 0.815 & 0.821 & 0.006 \\
(4)+(5) & 0.822 & 0.819 & 0.817 & 0.820 & 0.003 \\
(1)+(2)+(3)+(4)+(5) & 0.817 & 0.814 & 0.813 & 0.815 & 0.002 \\
(2)+(3)+(4)+(5) & 0.813 & 0.813 & 0.812 & 0.813 & 0.000 \\
(1)+(2)+(4)+(5) & 0.814 & 0.810 & 0.808 & 0.810 & 0.003 \\
(1)+(2)+(3)+(5) & 0.810 & 0.811 & 0.804 & 0.809 & 0.004 \\
(2)+(3)+(5) & 0.805 & 0.809 & 0.808 & 0.807 & 0.002 \\
(2)+(4)+(5) & 0.807 & 0.807 & 0.808 & 0.807 & 0.000 \\
(1)+(2)+(5) & 0.804 & 0.808 & 0.804 & 0.806 & 0.002 \\
(1)+(5) & 0.804 & 0.804 & 0.803 & 0.804 & 0.001 \\
(2)+(5) & 0.800 & 0.799 & 0.799 & 0.799 & 0.000 \\
\bottomrule
\end{tabular}
\caption{Accuracy of peer-reviewed reasoning under different judge prompt designs on the MedQA-USMLE dataset. We report results for 15 model combinations \added{over whole testset}. For each combination, Prompt 1–3 are evaluated, and Avg/Std are computed across the three prompts. \added{The relatively small std indicates that, although different prompts are used, the final performance varies only slightly. This suggests that our framework is robust to prompt variations.}}
\label{tab:prompt_acc}
\end{table*}

\subsection{Ablation study: biased peer-review}
Our method also includes cases where a model evaluates its own outputs. Although the model is not explicitly informed that the evaluated response is generated by itself, there remains a possibility that it may recognize its own answer and assign a higher score. To address this concern, we conduct additional experiments in which the judge models are different from the answer-generating model. For example, responses generated by the DeepSeek model are evaluated by the other four models as judges. As shown in Table \ref{tab:biased}, the biased peer-review is competitive to our method with slightly improvement.
\begin{table*}[htbp]
\centering
\begin{tabular}{l c c c c | c}
\toprule
Model & HeadQA(95\%CI) & MedQA(95\%CI) & PubMedQA(95\%CI) & Avg. & Unbiased\\
\midrule
(1)+(2)+(4) & 0.7582 (0.7049, 0.8115) & 0.6174 (0.5907, 0.6441) & 0.712 (0.672, 0.752) & 0.696 & 0.728\\
(2)+(3)+(4) & 0.7418 (0.6844, 0.7951) & 0.6112 (0.5844, 0.6379) & 0.764 (0.726, 0.8) & 0.706 & 0.737\\
(1)+(2)+(3)+(4) & 0.7418 (0.6844, 0.7951) & 0.6284 (0.6017, 0.6551) & 0.774 (0.736, 0.81) & 0.715 & 0.742\\
(1)+(3)+(4) & 0.7582 (0.7049, 0.8115) & 0.6952 (0.6701, 0.7203) & 0.802 (0.766, 0.836) & 0.752 & 0.761 \\
(1)+(2)+(5) & 0.8443 (0.7992, 0.8893) & 0.7989 (0.7769, 0.8209) & 0.728 (0.688, 0.766) & 0.790 & 0.791\\
(2)+(3)+(5) & 0.832 (0.7828, 0.877) & 0.802 (0.78, 0.824) & 0.746 (0.708, 0.784) & 0.793 & 0.795\\
(1)+(2)+(3)+(5) & 0.8361 (0.7869, 0.8811) & 0.806 (0.784, 0.8272) & 0.75 (0.712, 0.788) & 0.797 & 0.799\\
(2)+(4)+(5) & 0.8484 (0.8033, 0.8934) & 0.8068 (0.7848, 0.828) & 0.75 (0.712, 0.788) & 0.802 & 0.806\\
(1)+(2)+(4)+(5) & 0.8525 (0.8074, 0.8934) & 0.8091 (0.7871, 0.8303) & 0.75 (0.712, 0.788) & 0.804 & 0.809\\
(2)+(3)+(4)+(5) & 0.8443 (0.7992, 0.8893) & 0.8091 (0.7871, 0.8303) & 0.77 (0.732, 0.806) & 0.808 & 0.812\\
(1)+(4)+(5) & 0.8484 (0.8033, 0.8934) & 0.8225 (0.8013, 0.8429) & 0.756 (0.718, 0.794) & 0.809 & 0.813\\
(1)+(3)+(5) & 0.8484 (0.8033, 0.8934) & 0.8209 (0.7997, 0.8413) & 0.762 (0.724, 0.798) & 0.810 & 0.807\\
(1)+(2)+(3)+(4)+(5) & 0.8484 (0.8033, 0.8934) & 0.813 (0.791, 0.8342) & 0.77 (0.732, 0.806) & 0.810 & 0.814\\
(3)+(4)+(5) & 0.8525 (0.8074, 0.8934) & 0.8264 (0.8052, 0.8468) & 0.766 (0.728, 0.802) & 0.815 & 0.816 \\
(1)+(3)+(4)+(5) & 0.8566 (0.8115, 0.8975) & 0.828 (0.8068, 0.8484) & 0.778 (0.742, 0.814) & 0.821 & 0.820\\
\bottomrule
\end{tabular}
\caption{
Performance of biased peer-review settings across three medical QA datasets.
In this setting, the answer-generating model is excluded from the pool of judge models. 
Results are shown for the top 15 model combinations ranked by average accuracy under the biased peer-review setting; combinations may differ from those in Table \ref{tab:mainres}. Accuracy with 95\% confidence intervals is reported along with the average accuracy. The last column 'Unbiased' results are from the main experiments (Table \ref{tab:mainres}) for comparison
}
\label{tab:biased}
\end{table*}

\begin{figure}[t]
    \centering
    \includegraphics[width=0.8\linewidth]{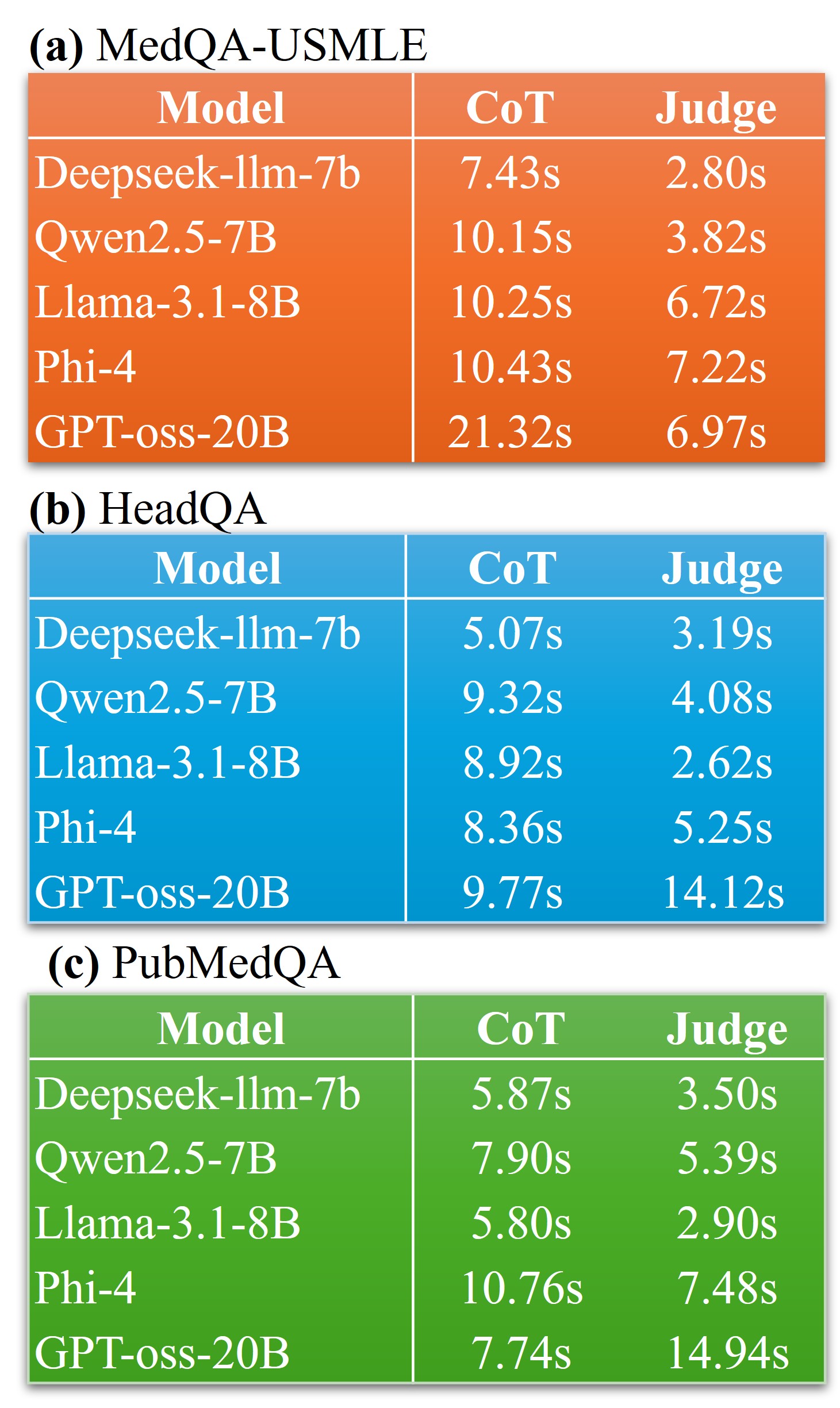}
    \caption{Runtime analysis of the proposed method across three datasets (MedQA-USMLE, HeadQA, and PubMedQA). For each model, average inference time is reported separately for Phase 1 (chain-of-thought generation) and Phase 2 (peer-review judgment). Overall, peer-review judgment is slightly faster than CoT generation, while the two-stage pipeline incurs higher total runtime than single-pass CoT prompting.}
    \label{fig:time}
\end{figure}
\subsection{Run time analysis}
Finally, we report the average runtime of our experiments across the three datasets, which consists of two phases: \textit{Phase 1}, CoT generation, and \textit{Phase 2}, peer-review judgment (as shown in the Fig. \ref{fig:time}). Overall, \textit{Phase 2} is slightly faster than \textit{Phase 1}, which is expected: the first phase requires generating a longer reasoning trajectory, whereas the second phase mainly focuses on scoring the peer-review outputs. Since our method follows a two-stage pipeline, it inevitably incurs higher computational cost and longer runtime than standard one-shot CoT prompting.

\section{Discussion}
MedQA is the fundamental task in biomedical natural language processing, and our proposed multi-agent method is designed for better question-answering ability from LLMs. 
Our experiments provide strong evidence that the proposed peer-reviewed reasoning method offers substantial improvements for MedQA. By systematically comparing against chain-of-thought reasoning and majority voting, we observed consistent gains in accuracy across HeadQA, MedQA-USMLE, and PubMedQA. Beyond accuracy, the peer review process successfully distinguished high- and low-quality reasoning chains, providing a novel way to enhance both robustness and interpretability in biomedical contexts.

The observed improvements can be attributed to the method’s focus on reasoning quality rather than final answers alone. Majority voting aggregates outputs from multiple models, but agreement does not guarantee correctness, as models trained on similar data often converge on the same flawed reasoning or biased predictions. Peer-reviewed reasoning addresses this limitation by incorporating cross-model evaluation of the reasoning process itself. Through structured peer assessments, logically sound reasoning chains are preferentially selected, even when they may not initially be in the majority. 
For example, while DeepSeek achieves lower accuracy as an individual solver, it introduces heterogeneity in the evaluation stage. As a judge, it penalizes certain flawed or shallow reasoning patterns that stronger models may occasionally overlook due to shared inductive biases. This diversity in judgment helps filter out low-quality reasoning chains, leading to improved overall performance despite DeepSeek’s weaker individual accuracy.
This mechanism mirrors the human peer-review process, in which arguments are evaluated not only by outcomes but also by their underlying validity.

Furthermore, our results highlight that peer-reviewed reasoning scales effectively with the number of participating models. As shown in Figure 2, performance improvements became more pronounced as additional agents were introduced. This trend suggests that 
differences in model outputs and reasoning tendencies across agents
provide complementary strengths, which can be systematically leveraged through peer evaluation. In practice, such scalability offers a pathway to stronger ensemble-style methods that avoid simple majority bias.

In biomedical applications, interpretability is not merely desirable but essential. Models deployed in clinical decision support or evidence synthesis must provide reasoning that clinicians and researchers can inspect. Traditional CoT reasoning improves transparency by revealing intermediate steps, but it does not inherently filter flawed reasoning. Peer-reviewed reasoning directly addresses this gap: the evaluative stage explicitly rewards coherent, medically sound reasoning chains. As demonstrated in Figure 3, LLM judges reliably differentiated strong from weak rationales, with high-performing models such as GPT-oss-20B and Phi-4 consistently receiving higher ratings. This evaluative dimension produces outputs that are both more accurate and more trustworthy for downstream use.

In addition, robustness is a critical requirement for biomedical AI. Peer review strengthens robustness by ensuring that final answers are supported by the most credible reasoning available in the agent pool. This mitigates risks of overfitting to spurious patterns or propagating systematic errors shared across models. By integrating both diversity (multiple agents) and selectivity (peer review), our method reduces variance in outcomes and enhances confidence in results.

Our method draws inspiration from two recent research directions: LLM-as-a-judge and multi-agent collaboration. Prior studies have shown that LLMs can achieve high agreement with human raters when evaluating biomedical relation extraction, summarization, and trial design outputs. Similarly, multi-agent approaches have demonstrated gains in medical consultation and diagnostic reasoning by encouraging collaboration among models. However, these directions have largely evolved separately. To our knowledge, this is the first work that explicitly integrates LLM-as-a-judge into a multi-agent reasoning paradigm for MedQA. By uniting evaluative and collaborative capacities, we show that LLMs can act as both solvers and reviewers, producing synergistic improvements that exceed either paradigm alone.

While our Peer Review method demonstrates significant performance gains over standard baselines, a critical consideration for practical deployment is the computational overhead. We acknowledge that the improved accuracy comes with increased resource consumption and wall-clock time. Our method involves three main stages: initial generation, peer review, and aggregation. The bottleneck lies in the peer review stage, which requires each of the $N$ agents to evaluate the $N$ reasonings produced, resulting in $N^2$ additional generation calls. Quantitatively, compared to the cost of a CoT generation (1 computational unit), the Majority Vote baseline with $N=5$ agents requires $N=5$ units. Our approach, however, necessitates $N + N^2 = 5 + 25 = 30$ computational units per question. This makes our method 5 times more resource-intensive than the majority vote baseline. Furthermore, due to the sequential dependency—the review stage must wait for all initial generations—the wall-clock time is also increased, approximately doubling that of the Majority Vote method, even when utilizing high parallelism for the review sub-tasks.

Despite these promising findings, several limitations should be acknowledged. First, the number of models and agent configurations was limited. While we observed consistent improvements up to five agents, scaling could reveal new challenges in coordination, computational cost, and aggregation strategy.
Second, due to the computation limitation, we didn't further explore the setting with more LLMs. With more models, the performance gain is not certain.
Third, we didn't consider the data leakage issues. The models we used might have been trained on these datasets.

Future research could focus on several extensions. One direction is to investigate integration with retrieval-augmented generation or external knowledge bases, which may further improve reasoning quality. In addition, expanding the method to multimodal inputs—such as combining text with imaging or laboratory data—would increase applicability to real-world clinical decision support.
Third, we acknowledge that tie handling represents a design choice that may influence final answer selection. While the simple deterministic strategy adopted here ensures reproducibility and consistency across methods, alternative tie-breaking mechanisms—such as secondary reasoning quality metrics, confidence estimation, or additional judging rounds—may further improve robustness. We leave systematic exploration of such strategies to future work.

Overall, our findings underscore the potential of peer-reviewed reasoning as a new paradigm for MedQA. By combining collaborative reasoning with evaluative judgment, our method enhances accuracy, interpretability, and robustness in ways that conventional methods cannot. As biomedical AI systems continue to move toward clinical deployment, approaches that emphasize both performance and trustworthiness will be essential. Peer-reviewed reasoning represents a promising step in this direction.

\section{Conclusion}
In this work, we proposed a multi-agent peer-reviewed reasoning method for medical question answering. By enabling large language models to act as both solvers and reviewers, the method consistently improved accuracy over chain-of-thought reasoning and majority voting across three benchmark datasets. Beyond higher accuracy, peer-reviewed reasoning enhanced interpretability and robustness by explicitly selecting reasoning chains of higher logical quality.

\section{Data Availability}0
The three datasets we used are all public:
\begin{enumerate}
    \item HeadQA: \href{https://aghie.github.io/head-qa/}{https://aghie.github.io/head-qa/}
    \item MedQA(USMLE): \href{https://github.com/jind11/MedQA}{https://github.com/jind11/MedQA}
    \item PubMedQA: \href{https://pubmedqa.github.io/}{https://pubmedqa.github.io/}
\end{enumerate}

\section{Code Availability}
The code is available at \href{https://github.com/Learner4everrr/Multi-agent-peer-reviewed-framework-for-MedQA}{https://github.com/Learner4everrr/Multi-agent-peer-reviewed-framework-for-MedQA}

\section{Acknowledgements}
We would like to express our sincere gratitude to the reviewers for their valuable comments.

\section{Author Contributions}
Z.Z.: Conceptualization, Data curation, Formal analysis, Investigation, Methodology, Validation, Visualization, Writing – original draft, Writing – review \& editing.
S.Z.: Validation, Writing – review \& editing.
R.Z.: Conceptualization, Project administration, Supervision, Writing – review \& editing.

\section{Competing Interests}
The authors declare no competing interests.

\section{Funding}
This work was supported by the National Institutes of Health’s National Center for Complementary and Integrative Health (grant numbers R01AT009457 and U01AT012871), the National Institute on Aging (grant number R01AG078154), the National Cancer Institute (grant number R01CA287413), the National Institute of Diabetes and Digestive and Kidney Diseases (grant number R01DK115629), and the National Institute on Minority Health and Health Disparities (grant number 1R21MD019134-01).

\bibliographystyle{unsrt}
\bibliography{0_main}

\end{document}